\documentclass[11pt]{article}

\usepackage{fullpage}
\usepackage{graphicx}
\usepackage{wrapfig}
\usepackage{cite}
\usepackage{hyperref}

\title{Past, Present, and Future Approaches Using Computer Vision for Animal Re-Identification from Camera Trap Data}
\author{Stefan Schneider \\ email \href{mailto:sschne01@uoguelph.ca}{sschne01@uoguelph.ca} 
   \and Graham W. Taylor \\ email \href{mailto:gwtaylor@uoguelph.ca}{gwtaylor@uoguelph.ca} 
   \and Stefan S. Linquist \\ email \href{mailto:linquist@uoguelph.ca}{linquist@uoguelph.ca} 
   \and Stefan C. Kremer \\ email \href{mailto:skremer@uoguelph.ca}{skremer@uoguelph.ca} }

\begin{document}
\maketitle

\textbf{Key Terms} - Animal Re-Identification, Camera Traps, Computer Vision, Convolutional Networks, Deep Learning, Density Estimation, Monitoring, Object Detection, Population Dynamics
\\
\newline
\noindent
\textbf{Author Details} - All four authors are based at the University of Guelph. Stefan Schneider is a PhD Candidate in the department of Computational Science and has ecological publications in The American Naturalist, Journal of Applied Ecology and the Conference on Computer and Robotic Vision. His thesis consists of developing an autonomous animal re-identification system powered by deep learning and applying this methodology to octopus ethology. Graham Taylor is an associate professor of Engineering and a member of the Vector Institute for Artificial Intelligence. He is an internationally-recognized expert on Deep Learning and has published at conferences such as Neural Information Processing Systems, International Conference on Machine Learning, and journals such as IEEE Transactions on Pattern Analysis and Machine Intelligence and the Journal of Machine Learning Research. Stefan Linquist is an associate professor in Philosophy and Integrative Biology. His articles on ecological methods appear in Molecular Ecology, Biological Reviews, and The American Naturalist. Stefan C. Kremer is professor of Computer Science. He specializes in machine learning and bioinformatics and has published papers in the Journal of Machine Learning Research, Neural Computation, IEEE Transactions on Computational Biology and Bioinformatics, Genome, and BNC Bioinformatics. 
\begin{center}
Word Count - 6,996
\end{center}
\newpage

\section*{Abstract}
1. The ability of a researcher to re-identify (re-ID) an individual animal upon re-encounter is fundamental for addressing a broad range of questions in the study of ecosystem function, community and population dynamics, and behavioural ecology. Tagging animals during mark and recapture studies is the most common method for reliable animal re-ID however camera traps are a desirable alternative, requiring less labour, much less intrusion, and prolonged and continuous monitoring into an environment. Despite these advantages, the analyses of camera traps and video for re-ID by humans are criticized for their biases related to human judgment and inconsistencies between analyses.
\newline
2. In this review, we describe a brief history of camera traps for re-ID, present a collection of computer vision feature engineering methodologies previously used for animal re-ID, provide an introduction to the underlying mechanisms of deep learning relevant to animal re-ID, highlight the success of deep learning methods for human re-ID, describe the few ecological studies currently utilizing deep learning for camera trap analyses, and our predictions for near future methodologies based on the rapid development of deep learning methods.
\newline
3. For decades ecologists with expertise in computer vision have successfully utilized feature engineering to extract meaningful features from camera trap images to improve the statistical rigor of individual comparisons and remove human bias from their camera trap analyses. Recent years have witnessed the emergence of deep learning systems which have demonstrated the accurate re-ID of humans based on image and video data with near perfect accuracy. Despite this success, ecologists have yet to utilize these approaches for animal re-ID.
\newline
4. By utilizing novel deep learning methods for object detection and similarity comparisons, ecologists can extract animals from an image/video data and train deep learning classifiers to re-ID animal individuals beyond the capabilities of a human observer. This methodology will allow ecologists with camera/video trap data to re-identify individuals that exit and re-enter the camera frame. Our expectation is that this is just the beginning of a major trend that could stand to revolutionize the analysis of camera trap data and, ultimately, our approach to animal ecology.

\newpage

\section*{Introduction}

The ability to re-ID animals allows for population estimates which are used in a variety of ecological metrics including diversity, evenness, richness, relative abundance distribution, and carrying capacity, which contribute to larger, overarching ecological interpretations of trophic interactions and population dynamics \cite{krebs1989ecological}. Ecologists have used a variety of techniques for re-ID including tagging, scarring, banding, and DNA analyses of hair follicles or feces \cite{krebs1989ecological}. While accurate, these techniques are laborious for the field research team, intrusive to the animal, and often expensive for the researcher.
\newline
\\
Compared to traditional methods of field observations, camera traps are desirable due to their lower cost and reduced workload for field researchers. Camera traps also provide a unique advantage by recording the undisturbed behaviours of animals within their environment. This has resulted in the discovery of surprising ecological interactions, habitat ranges, and social dynamics, among other insights \cite{scheel2017second, meek2013reliability}. These advantages have led to a 50\% annual growth in publications using camera trap methods to assess population sizes between 1998 and 2008 and the trend has persisted until 2015 \cite{rowcliffe2008estimating, burton2015wildlife}. 
\newline
\\
Despite their advantages, there are a number of practical and methodological challenges associated with the use of camera traps for animal re-ID. The discrimination of individual animals is often an expert skill requiring a considerable amount of training. Even among experienced researchers there remains an opportunity for human error and bias \cite{foster2012critique, meek2013reliability}. Historically, these limitations have restricted the use of camera traps to the re-ID of animals that bear conspicuous individual markings. 
\newline
\\
One strategy for overcoming these limitations has involved the use of computer vision to standardize the statistical analysis of animal re-ID. For decades `feature engineering' has been the most commonly used computational technique where algorithms are designed and implemented to focus exclusively on pre-determined traits, such as patterns of spots or stripes, to discriminate among individuals. The main limitations of this approach surround its impracticality \cite{hiby2009tiger}. Feature engineering requires programming experience, sufficient familiarity with the organisms to identify relevant features, and lacks in generality where once a feature detection algorithm has been designed for one species, it is unlikely to be useful for other taxa.  
\newline
\\ 
Recent decades have witnessed the emergence of deep learning systems that make use of large data volumes \cite{zheng2015scalable}. Modern deep learning systems no longer require `hard-coded' feature extraction methods. Instead, these algorithms can learn, through their exposure to large amounts of data, the particular features that allow for the discrimination of individuals. \cite{lecun2015deep}. These methods have been developed primarily outside the realm of ecology, first in the field of computer image recognition \cite{krizhevsky2012imagenet}, and more recently in the security and social media industries \cite{zheng2015scalable}. Modern deep learning systems now consistently outperform feature engineered methods provided that they have access to large amounts of data \cite{lisanti2015person, martinel2015re}. 
\newline
\\
In recent years, a handful of ecologists have begun utilizing deep learning systems for species and animal individual identification with great success \cite{norouzzadehautomatically, schneider2018deep, carter2014automated, freytag2016chimpanzee, brust2017towards, loos2013automated}. Our expectation is that this is just the beginning of a major trend that could stand to revolutionize the analysis of camera trap data and, ultimately, our approach to animal ecology. Here we present a collection of works utilizing computer vision for animal re-ID. We begin with the earliest computer aided and featured engineered approaches followed by an introduction to deep learning relevant to animal re-ID. We then present recent works utilizing deep learning for animal re-ID and lastly conclude discussing its practical applications. 

\section*{Computer Vision Feature Extraction Methods for Animal Re-Identification}

A common metric when reporting results classification results is top-1 accuracy which describes what percentage of queries from an unseen test set the model answered correctly. An overall summary of the reviewed studies can be found in Figure 1, Table 1, \& Appendix A.
\newline
\\
The first use of computer vision for animal re-ID was introduced in 1990 by Whitehead, Mizroch et al., and Hiby and Lovell who published collectively in the same journal \cite{whitehead1990computer, mizroch1990computer, hiby1990computer}. Whitehead (1990) considered sperm whales (\textit{Physeter macrocephalus}) using custom software to scan projector slides of a sperm whales fluke onto a digitizer tablet \cite{whitehead1990computer}. The user would manually tap and save the location of a unique characteristic (such as a nick or a scratch) to a database. The software then considers the maximum sum of similarities of the descriptors and return the most similar individual. Considering images collected from the Galapagos Islands, Ecuador containing 1,015 images of all unique individuals, 56 were considered for testing where the approached returned a 32\% top-1 accuracy \cite{whitehead1990computer}. Mizroch et al.\ (1990) considered humpback whales (\textit{Megaptera novaeangliae}) re-identification using customized software where the user labeled images of whale flukes considering a 14-sector fluke map selecting from a list of possible markings (Spots, pigment, etc.). Using a collection of 9,051 images of 790 individuals provided by the National Marine Mammal Laboratory a similarity matrix was used to compare the ID of 30 individuals returning a top-1 accuracy of 43\% \cite{mizroch1990computer}. Re-ID from images of a whale's fluke remains a research interest today \cite{kagglehumpbackreid}. These earliest approaches rely on qualitative descriptors which overall are too limited of a representation to capturing the nuanced detail required for animal re-ID. In 1990, Hiby and Lovell introduced the first feature engineered system considering the grey seal (\textit{Halichoerus grypus}). The user inputs numerous head captures of an animal from the same pose and the system renders a grid based 3-D model of the animal where an individual's representation is captured in the summation of the grey scale pixel values of each grid cell \cite{hiby1990computer}. Considering a created data set using photos taken from an undisclosed beach during grey seal breeding, a test set of 58 images of 58 individuals were re-identified with a 98\% top-1 accuracy \cite{hiby1990computer}. This technique is limited to animals being in a specific orientation for the photos. In 1998, O'Corry-Crowe considered a feature engineering approach using a wavelet transformation to numerically represent the fluke of sperm whales using images provided by the Andenes Whale Center in Norway receiving a 92.0\% accuracy considering 56 images of 8 individuals \cite{o1998identification}. The improved performance and unbiased results of these two feature extraction methods demonstrated their future potential for animal re-ID.
\newline
\\
In the 2000s Kelly (2001) used the same 3-D model approach as Hiby and Lovell (1990), comparing the collective similarity of pattern cells of cheetahs (\textit{Acinonyx jubatus}) \cite{kelly2001computer, hiby1990computer}. Considering a catalogue of 10,000 images of from the Serengeti National Park in Tanzania, Kelly (2001) achieved a 97.5\% top-1 accuracy using a test set of 1,000 images of an undisclosed number of individuals \cite{kelly2001computer}. In 2003, Hillman et al., developed an autonomous feature extraction method to capture unique nicks, scratches, and markings from the dorsal fin of six whale and dolphin species \cite{hillman2003computer}. Named \textit{Finscan}, the system localized 300 xy pairs from the dorsal fin of a variety of dolphin and shark species and determines individual similarity by calculating the minimum Euclidean distance of these pairs \cite{hillman2003computer}. Hillman et al. (2003) report accurate classification results of 50\% top-1 \cite{hillman2003computer} on their own data sets containing \~500 images with on average 52 images of 36 individuals per species. This technique is ultimately limited by compressing an animal's representation into a single numeric value.
\newline
\\
The complexity of feature representation increased in 2004 as Ravela and Gamble used a Taylor approximation of local colour intensities, multi-scale brightness histograms, and curvature/orientation to re-ID marbled salamander (\textit{Ambystoma opacum}) individuals  \cite{ravela2004recognizing}. Ravela and Gamble (2004) constructed a custom housing for their camera trap to capture top down images of salamanders as they entered. Using their collected images from traps in Massachusetts, USA a database of 370 individuals was gathered and considering a test set of 69 images with an undisclosed number of individuals, they returned a top-1 accuracy of 72\% \cite{ravela2004recognizing}. In 2005, Arzoumanian et al.\ explored whale shark (\textit{Rhincodon typus}) re-ID by analyzing the characteristic flank (front dorsal region) spot patterns using an astronomy algorithm for recognizing star patterns \cite{groth1986pattern, arzoumanian2005astronomical}. The algorithm creates a list of possible triangles from white dots within an image using xy pairs and compares their orientation between images. Arzoumanian et al.\ (2005) used the ECOCEAN Whale Shark Photo-Identification library of 271 `datasets' to test 27 images of individuals from an undisclosed number of classifications to report a top-1 accuracy of 90\% \cite{arzoumanian2005astronomical}. In 2007, Ardovini et al.\ attempted to re-identify elephants (\textit{Loxodonta spp.}) based on images using a multi-curve matching technique where a spline curve is fit to the mantle of an individual elephant \cite{ardovini2008identifying}. To re-ID an elephant, the known spline curves were overlaid atop the image of the elephant and the most similar was considered the same individual \cite{ardovini2008identifying}. Considering a database of elephants from Zakouma Ciad National Park consisting of 268 individuals and 332 test images, the researchers report a 75\% top-1 accuracy. The methodology has since been used by the Centre for African Conservation \cite{ardovini2008identifying}. Overall, the studies so far are limited by their ability to capture nuanced details required for animal re-ID. 
\newline
\\
The first computer vision model for animal re-ID capable of generalizing across species was developed in 2007 by Burghardt and Campbell which extracted inherent singularities (spots, lines, etc.) of animal individuals \cite{burghardt2007fully}. The technique involves three cameras pointed at a central location to capture a 3-D representation of the animal and extracts features using Haar-like (pixel difference) descriptors. For each feature, an AdaBoost classifier was trained to determine if the single feature was present and individual re-ID governed by an ensemble of these features \cite{burghardt2007fully}. In 2010, Sherley et al. used this approach on Robben Island, Africa in one of the first works to perform fully autonomous population estimates solely from camera traps considering the African penguin (\textit{Spheniscus demersus}) \cite{sherley2010spotting}. The system returned an accuracy between 92-97\% in comparison to records from humans on site \cite{sherley2010spotting}. Overall this method is limited by the labourious task of training unique classifier for each possible feature on an animal. 
\newline
\\
In 2009, Hiby et al. collaborated with Karanath to explore their 3-D model representation technique considering a database of images from his long term tiger population studies in Nagarhole and Bandipur tiger reserves \cite{hiby2009tiger}. By again dividing individuals into pattern cells, re-ID was considered by the summation of the similarity of the cells were considered using a Bayesian posterior probability estimate and tested considering live tigers and poached tiger rugs. Hiby et al.\ (2009) report a 95\% top-1 accuracy considering 298 individuals from 298 possible individuals \cite{hiby2009tiger}. 
\newline
\\
In 2013, Town et al.\ compared individual similarity of manta ray species (\textit{Manta alfredi} and \textit{Manta birostris}) based on images of their ventral side enhancing local contrasts and then extracting features (ie. spots) using the SIFT algorithm \cite{town2013manta}. Considering a test size of 720 images of 265 individuals from the Manta Ray \& Whale Shark Research Centre Inhambane, Mozambique, Town et al. report a 51\% accuracies commenting on the limitations of underwater photography \cite{town2013manta}. Also in 2013, Loos and Ernst attempt to re-ID the faces of chimpanzee (\textit{Pan spp.}) considering features of pixel value gradient changes and pixel groupings to train a Support Vector Machine classifier, a classifier which maximizes a linear decision boundary between classifications \cite{loos2013automated, hearst1998support}. Loos and Ernst (2013) report an accuracy of 84.0\% and 68.8\% on their two labeled data sets, C-Zoo from the Leipzig Zoo, Germany containing 598 images of 24 individuals and C-Tai from the Tai National Park, Cote d'Ivoire, Africa containing 1,432 images of 71 individuals, using a randomly selected 1/5th of the data as test images \cite{loos2013automated}. Loos and Ernst (2013) consider a promising approach but are limited by the capabilities of the SVM classifier. In 2017, Hughes and Burghardt use fin contours to extract features and use a local naive Bayes nearest neighbour non-linear model for re-ID \cite{hughes2017automated}. Using the data set FinsScholl2456 containing 2456 images of 85 individuals, their approach returns a top-1 accuracy of 82\% for correctly classified shark individuals demonstrating strong performance for re-ID by dorsal fin.
\newline
\\
While these techniques have shown success, there remains improvement in the performance, ease of design, implementation and overall accessibility of these algorithms. Deep learning methods provide an opportunity to remedy each of these disadvantages. 

\section*{Deep Learning and Its Success for Human Re-Identification}

Modern deep learning systems have shown great success learning the necessary features for re-ID from data and removes the need for feature engineering. Deep learning as concept originated in the 1940-1960s as cybernetics, was rebranded as connectionism between 1980 and 1995, and starting in 2006 rebranded again as deep learning \cite{mcculloch1943logical, rumelhart1986learning, hinton2006fast}. Despite its long history, there has been a rapid growth of interest in deep learning due to its success related to improved computational power and the availability of large data sets, both requirements for the model. In recent years, deep learning methods have dramatically improved performance levels in the fields of speech recognition, computer vision, drug discovery, genomics, artifical intelligence, and others becoming the standard computational approach for problems with large amounts of data \cite{lecun2015deep}. Here, we provide a brief description of the underlying mechanism of deep learning as it relates to computer vision and animal re-ID. The process that will be outlined in this section is an approach known as supervised learning and is the most common form of deep learning \cite{lecun2015deep}. 
\newline
\\
Deep learning is a computational framework where a system's parameters are not designed by human engineers but trained from large amounts of data using a general-purpose learning algorithm \cite{lecun2015deep}. Intuitively, deep learning systems are organized as a layered web/graph structure where labeled data are submitted as input, and many modifiable scalar variables (known as weights) are summed and multiplied by a non-linear transformations to output a predicted label from a predefined number of possible choices \cite{goodfellow2016deep}. Each training example is fed through the network providing an answer, and based on the results from an `objective function', (ie. the average answer accuracy across the data set) the weight values are modified by an optimizer (e.g. gradient descent with backpropagation) in an attempt to improve accuracy \cite{goodfellow2016deep}. Deep learning systems will often have millions of modifiable weight values and with enough data and computation, the underlying relationship between the data and output can be mapped to return accurate results \cite{krizhevsky2012imagenet, simonyan2014very, szegedy2015going, he2016deep}. The general term for this framework is a neural network of which there are many architectures. The architecture described here is known as a feedfoward network \cite{lecun2015deep}. In 1991, Hornik proved feedforward neural networks are a universal approximator, capable of mapping any input to output if a relationship exists \cite{hornik1991approximation}.
\newline
\\
Neural networks are best described as multiple processing layers where each layer learns a representation of the data with different levels of abstraction \cite{lecun2015deep}. As the data representation passes through each transformation, it allows for complex representations to be learned \cite{lecun2015deep}. Consider an example data set of many black and white images of animal individuals. The images are initially unravelled into a single vector of pixel values between 0 and 255 and fed into a deep learning system. Using this raw input the first layer is able to learn simple representations of patterns within the data, such as the presence or absence of edges at particular orientations and locations in the image. The second layer will typically represent particular arrangements of edges and open space. The third layer may assemble these edges into larger combinations that correspond to parts of familiar objects, such as the basics of a nose, or eyes, and subsequent layers would detect objects as combinations of these parts, such as a face \cite{lecun2015deep}. Based on the combination of larger parts, such as the ears, face, or nose, a learning system is able to correctly classify different individuals with near perfect accuracy when given enough input data \cite{swanson2015snapshot}. A deep learning system can learn the subtle details distinguishing a unique classification, such as a differing antler structure, and ignore large irrelevant variations such as the background, pose, lightning, and surrounding objects \cite{krizhevsky2012imagenet}.
\newline
\\
Many recent advances in machine learning have come from improving the architectures of a neural network. One such architecture is the Convolutional Neural Network (CNN), first introduced as the `Neocognitron' in 1979 by Fukushima, which is now the most commonly used architecture for computer vision tasks today \cite{fukushima1979neural, krizhevsky2012imagenet}. CNNs introduce `convolutional' layers within a network which learn feature maps that represent the spatial similarity of patterns found within the image (e.g. the presence or absence of lines or colours within an image) \cite{lecun2015deep}. Each feature map is governed by a set of `filter banks', which are matrices of scalar values that are learned similar to the standard weights of a network. Intuitively, these feature maps learn to extract the necessary features for a classification task replacing the need for feature engineering. CNNs also introduce max pooling layers, a method that reduces computation and increases robustness by evenly dividing the feature map into regions and returning only the highest activation values \cite{lecun2015deep}. A simple CNN will have a two or three convolution layers passed through non-linearity functions, interspersed with two or three pooling layers, ending with fully-connected layers to return a classification output. Machine learning researchers continually experiment with modular architectures of neural networks and six CNN frameworks have standardized as well-performing with differences generally considering computation cost/memory in comparison to accuracy. These networks include AlexNet, VGG16, GoogLeNet/InceptionNet, ResNet, DenseNet and CapsNet \cite{krizhevsky2012imagenet, simonyan2014very, szegedy2015going, he2016deep, iandola2014densenet, sabour2017dynamic}. These networks range from 7 to 152 layers. A common approach for training a network for a niche task like animal re-ID is to use the pre-trained weights of one of these six network structures trained on a public data set as initialization parameters, and then retraining the network using labeled data of animal individuals \cite{pan2010survey}. This is approach is known as Transfer Learning and helps improve performance when training on limited amounts of data\cite{pan2010survey}.
\newline
\\
One niche task of deep learning research focuses on improving performance on extremely similar classifications, known as fine-grained visual recognition. Species identification tasks, such as classifying 675 similar moth species, has been an active research area for testing this problem domain and applicable for animal re-ID \cite{rodner2015fine}. Two techniques have been demonstrated to improve performance. The first is data augmentation, a universally recommended approach when training computer vision networks. This involves randomly flipping, cropping, rotating, blurring, shifting, altering colour/light images during each iteration of training \cite{rodner2016fine}. A second is an approach known as `object localization', where classification predictions are made for each unique parts of the body of an animal (ie. head, beck, ears, wings, etc.) that are then all considered in combination for a final classification \cite{souri2015fast}.
\newline
\\
The deep learning methods described so far are limited to returning one animal classification per image, however this is suboptimal for the analysis of camera trap images. In order to identify multiple objects (ie. animals), researchers train object detectors which segregate the image into regions that are passed through a CNN. Three approaches for object detection have recently grown in popularity. The first is Faster Region-CNN which segregates the image into approximately 2000 proposal regions and passes each through a CNN \cite{ren2017faster}. The second is YOLO (You-Only-Look-Once) which divides an image into a grid, and passes each grid cell through the network considering a series of predefined `anchors' relevant to the predicted shape and size classifications of interest \cite{redmon2016you}. Lastly is Single Shot Multibox Detector, which considers a set of default boxes and makes adjustments to these boxes during training to align over the objects of interest \cite{liu2016ssd}. Object detection methods have an additional objective function evaluation metric known as Intersection over Union (IOU), which returns performance as the area of overlap of the true and predicted regions divided by the entire area of the true and predicted regions \cite{nowozin2014optimal}. 
\newline
\\
In 2015, two research teams, Lisanti et al. and Martinel et al., demonstrated the successful capabilities of CNNs on human re-ID using the ETHZ data set, a data set composed of 8580 images of 148 unique individuals taken from mobile platforms, where CNNs were able to correctly classify individuals from the test set with 99.9\% accuracy after seeing 5 images of an individual \cite{lisanti2015person, martinel2015re}. In 2014, Taigman et al. introduced Deepface, a method of creating a 3-dimensional representation of the human face to provide more data to a neural network which returned an accuracy of 91.4\% on the YouTube faces dataset containing videos of 1,595 individuals \cite{taigman2014deepface}. Considering traditional CNNs for re-ID requires a large number of labeled data for each individual and re-training the network for every new individual sighted, both of which are infeasible requirement for animal re-ID. In 1993, Bromley et al. introduced a suitable neural network architecture for this problem, titled a Siamese network, which learns to detect if two input images are similar or dissimilar \cite{bromley1994signature}. Once trained, Siamese networks require only one labeled input image of an individual in order to accurately re-identify if an individual in a second image. In practice, one would train a Siamese network to learn a species' similarity and compare individuals from a known database. In 2016, Schroff et al. introduced FaceNet which uses a three image Siamese image to train similarity using an input image as well as a similar and dissimilar image \cite{schroff2015facenet}. FaceNet currently holds the highest accuracy on the YouTube Faces data set with a 95.12\% top-1 accuracy \cite{schroff2015facenet}. 
\newline
\\
Based on the results for human re-ID, deep learning systems show promise as a generalizable framework for animal re-ID eliminating the biases related to human judgment and reducing the requirement for human labour to analyze images. With enough data, deep learning systems can be trained for any species which eliminates the need for domain expertise and species specific feature extraction algorithms. Deep learning systems also show promise to exceed human level performance when re-identifying animal individuals without obvious patterns and markings. 

\section*{Deep Learning for Camera Trap Analysis of Species and Animal Re-Identification}

Despite the success of deep learning methods for human re-ID, few ecological studies have utilized its advantages. In 2014, Carter et al.\ published one of the first works using neural networks for animal re-ID, a tool for green turtle (\textit{Chelonia mydas}) re-ID \cite{carter2014automated}. The authors collected 180 photos of 72 individuals from Lady Elliot Island in the southern Great Barrier Reef, both nesting and free swimming considering an undisclosed number of testing images. Their algorithm pre-processes the image by extracting a shell pattern, converting it to grey scale, unravelling the data into a raw input vector, and then training a simple feedforward network \cite{carter2014automated}. Each model produces an output accuracy of 80-85\% accuracy, but the authors utilize an ensemble approach by training 50 different networks and having each vote for a correct classification. The ensemble approach returns an accuracy of 95\%. Carter et al.'s work has been considered a large success and is currently used to monitor the southern Great Barrier Reef green turtle population \cite{carter2014automated}. 
\newline
\\
In 2016, Freytag et al.\ trained the CNN architecture AlexNet on the isolated faces of chimpanzees considering the C-Zoo and C-Tai data sets \cite{freytag2016chimpanzee}. Freytag et al.\ (2016) report an improved accuracy of 92.0\% and 75.7\% in comparison to the original feature extraction method of 84.0\% and 68.8\% \cite{freytag2016chimpanzee, loos2013automated}. In 2017, Brust et al.\ trained the object detection method YOLO to extract cropped images of Gorilla \textit{Gorilla gorilla} faces from 2,500 annotated images camera trap images of 482 individuals taken in the Western Lowlands of the Nouabal\'e -Nodki National Park in the Republic of Congo \cite{brust2017towards}. Once the faces are extracted, Brust et al.\ (2017) followed the same procedure as Freytag et al.\ (2016) to train the CNN AlexNet achieving a 90.8\% accuracy on a test size of 500 images \cite{freytag2016chimpanzee, brust2017towards}. The authors close discussing how deep learning for ecological studies show promises for a whole realm of new applications if the fields of basic identify, spatio-temporal coverage and socio-ecological insights. \cite{freytag2016chimpanzee, brust2017towards}
\newline
\\
In 2018, Koerschens considered a completed pipeline of object detector and classifer to re-identify elephant individuals considering a dataset from \textit{The Elephant Listening Project} from the Dzanga-Sangha reserve of which 2,078 images of 276 different individuals are present \cite{Koerschens18:Elephants}. Koerschens use a YOLO object detector trained using the ResNet50 architecture to localize the head of an elephant from an image, and pass each head through a train Support Vector Machine classifier \cite{Koerschens18:Elephants}. Their model returns a top-1 accuracy of 59\% \cite{Koerschens18:Elephants}. Most recently, Deb et al. (2018) tested the re-ID capabilities of deep learning systems on three primate spieces: chimpanzees, lemurs (\textit{Lemuriformes spp.}), and golden monkeys (\textit{Cercopithecus kandti}) \cite{deb2018face}. Deb et al. (2018) consider three metrics for re-ID: verification (two image comparison), closed-set identification (individual is known to exist within the data), open-set identification (individual may or may not exist within the data) \cite{deb2018face}. For chimpanzees, Deb et al. (2018) combined the C-Zoo and C-Tai data sets to create the \textit{ChimpFace} data set containing 5,599 images of 90 chimpanzees. For lemurs, they consider a data set known as \textit{LemurFace} from the Duke Lemer Center, North Carolina containing 3,000 face images of 129 lemur individuals from 12 different species. For golden monkeys, they extracted the faces of 241 short video clips (average 6 seconds) from Volcanoes National Park in Rwanda where 1,450 images of 49 golden monkey faces were cropped and extracted \cite{deb2018face}. Deb et al., (2018) use a custom Siamese CNN containing four convolutional layers, followed by a 512 node fully connected layer \cite{deb2018face}. Deb et al. (2018) report verification, closed-set, and open-set accuracies respectively for lemurs: 83.1\%, 93.8\%, 81.3\%, golden monkeys: 78.7\%, 90.4\%, 66.1\%, and chimpanzess: 59.9\%, 75.8\%, and 37.1\% \cite{deb2018face}. 

\section*{Near Future Techniques for Animal Re-Identification}
Ecologists familiar with techniques in computer vision have exhibited considerable ingenuity in developing feature extraction methods for animal re-ID, but only recently have researchers considered modern deep learning methods, such as CNNs or Siamese networks. By considering modern deep learning approaches, ecologists can utilize improve accuracies without the requirement of hand-coded feature extraction methods by training a neural network from large amounts of data. 
\newline
\\
We foresee the greatest challenge for deep learning methods being the creation of large labeled datasets for animal individuals. From our review, we suggest 1,000+ images are required to train deep networks. Our proposed approach for data collection would be to utilize environments with known ground truths for individuals, such as national parks, zoos, or camera traps in combination with individuals being tracked by GPS, to build the datasets. We would recommend using video wherever possible to gather the greatest number of images for a given encounter with an individual. We encourage researchers with images of labeled animal individuals to make these datasets publicly available to further the research in this field. In addition to gathering the images, labeling the data is then also a labourious task, especially when training an object detection model where bounding boxes are required. One approach for solving this problem is known as weakly supervised learning, where one provides object labels to a network (ie. zebra) and the network returns the predicted coordinates of its location \cite{zhou2017brief}. An alternative approach is to outsource the labeling task to online services, such as Zooniverse which can be time saving for researchers, but introduces inevitable error and variability \cite{simpson2014zooniverse}. 
\newline
\\
While deep learning approaches are able to generalize to examples similar to those seen during training, we foresee various environmental, positional, and timing related challenges. Environmental difficulties may include challenging weather conditions, such as heavy rain, or extreme lighting/shadows. A possible solution to limit these concerns may be to re-ID only during optimal weather conditions. A positional challenge may occur if an individual were to enter the camera frame at extremely near or far distances. To solve this, one could limit animals to a certain range from the camera before considering it for re-ID. Lastly, a challenge may be if an individual's appearance were to change dramatically between sightings, such as being injured or the rapid growth of a youth. While a network would be robust to such changes given training examples, this would require examples be available as training data. To account for this issue we would consider having a `human-in-the-loop' approach, where a human monitors results and relabels errorenous classifications for further training to improve performance \cite{holzinger2016interactive}. 
\newline
\\
While today fully autonomous re-ID is still in development, researchers can already use these systems to reduce manual labour for their studies. Examples include training networks to filter images by the presence/absence of animals, or species classifications \cite{kaggleiWildCam2018, norouzzadehautomatically, schneider2018deep}. Soon deep learning systems will accurately perform animal re-ID at which time one can create systems that autonomously extract from camera traps a variety of ecological metrics such as diversity, evenness, richness, relative abundance distribution, carrying capacity, and trophic function, contributing to overarching ecological interpretations of trophic interactions and population dynamics.

\section*{Conclusion}
Population estimates are the underlying metric for many fundamental ecological questions and rely on the accurate re-identification of animals. Camera and video data have become increasingly common due to their relatively inexpensive method of data collection, however they are criticized for their unreliability and bias towards animals with obvious markings. Feature engineering methods for computer vision have shown success re-identifying animal individuals and removing biases from these analyses, however these methods require algorithms designed for feature extraction. Deep learning provides a promising alternative for ecologists as it learns these features from large amounts and has shown success for human and animal re-ID. By utilizing deep learning methods for object detection and similarity comparison, ecologists can utilize deep learning methods to autonomously re-identify animal individuals from camera trap data. Such a tool would allow ecologists to automate population estimates.

\newpage
\begin{table}[h!]
\centering
\caption{Summary of Feature Engineered and Deep Learning Approaches for Animal Re-ID.}
\begin{tabular}{ l c l c c c}
	\multicolumn{6}{ c }{Computer Vision Animal Re-Identification Techniques}\\
	\hline
	Animal & Year & \centering{Methodology} & Test Size & \multicolumn{1}{p{1.5cm}}{\centering Num. \\ Classes} & \multicolumn{1}{p{2.3cm}}{\centering Top-1 \\ Accuracy (\%)} \\ \hline
	Sperm Whale & 1990 & Database similarity & 56 & 1,015& 59 \\
	Humpback Whale & 1990 & Database similarity & 30 & 790 & 41.4 \\
	Grey Seal & 1990 & 3-D Pattern Cell similarity & 58 & 58 & 98.0 \\
	Sperm Whale & 1998 & Wavelet transformations & 56 & 8 & 92.0 \\
	Cheetah & 2001 & 3-D Pattern Cell similarity & 1,000 & NA & 97.5 \\
	Whale/Dolphin & 2003 & XY Pair Euclidean Distance & 52 & 36 & 50.0 \\
	Marbled Salamander & 2004 & Pixel histogram and local colours & 69 & NA & 72.0 \\
	Whale Shark & 2005 & Star pattern recognition & 27 & NA & 90.0 \\
	Elephant & 2007 & Polynomial multi-curve matching & 332 & 268 & 75.0 \\
	African penguin & 2009 & Per feature AdaBoost classifer & N/A & NA & 92-97.0 \\
	Tiger & 2009 & 3-D Pattern Cell similarity & 298 & 298 & 95.0 \\
	Manta Ray & 2013 & SIFT & 720 & 265 & 51.0 \\
	Chimpanzee (C-Zoo) & 2013 & Support Vector Machine & 478 & 120 & 84.0 \\
	Chimpanzee (C-Tai) & 2013 & Support Vector Machine & 1146 & 286 & 68.8 \\
	Green Turtule & 2014 & Feedforward Network & 180 & 72 & 95.0 \\
	Chimpanzee (C-Zoo) & 2016 & Convolutional Network & 478 & 120 & 92.0 \\
	Chimpanzee (C-Tai) & 2016 & Convolutional Network & 1146 & 286 & 75.7 \\
	Shark & 2017 & Naive Bayes Nearest Neighbour & 2456 & 85 & 82.0 \\
	Gorilla & 2017 & Convolutional Network & 500 & 482 & 90.8 \\
	Elephant & 2018 & Support Vector Machine & 2,078 & 276 & 59.0 \\
	Chimpanzee & 2018 & Siamese Network & 5,599 & 90 & 93.8 \\
	Lemur & 2018 & Siamese Network & 3,000 & 129 & 90.4 \\
	Golden Monkey & 2018 & Siamese Network & 241 videos & 49 & 75.8 \\
\end{tabular}
\end{table}

\begin{figure}[t!]
	\centering
	\textbf{Time Line of Computer Assisted Approaches}\par
	\includegraphics[width=17cm]{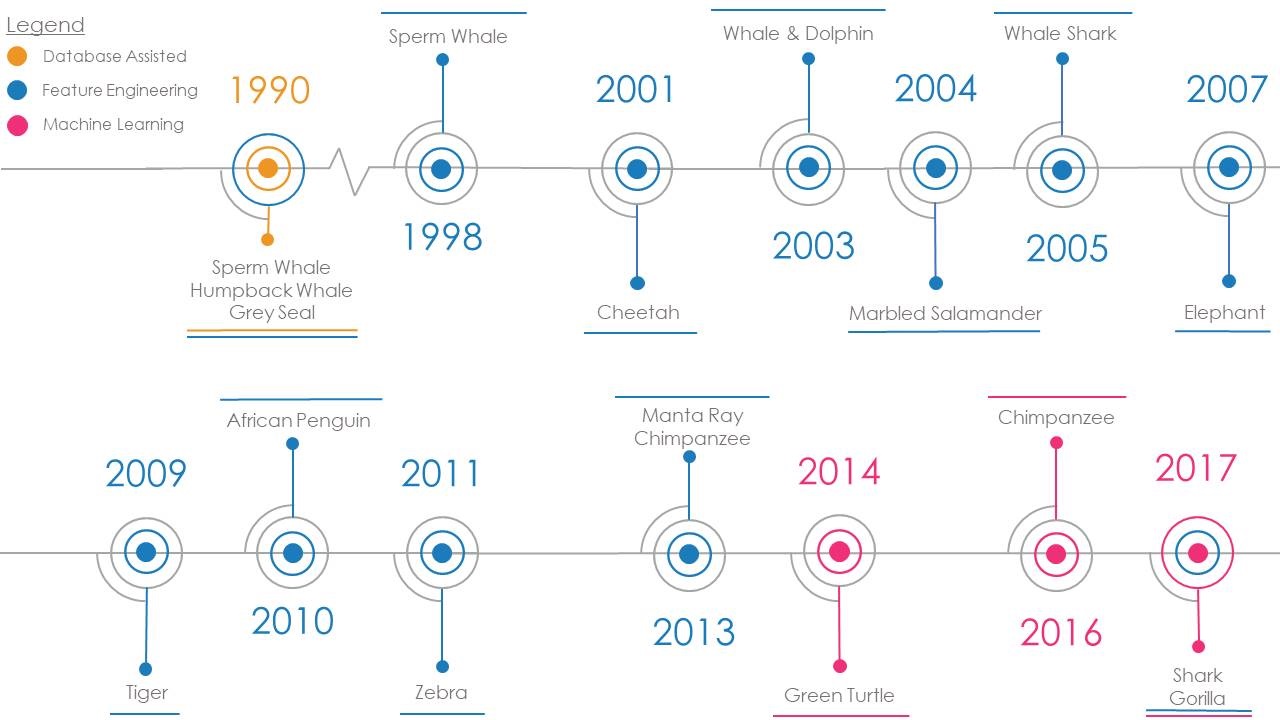}
	\caption{Time line of animal re-ID methods discussed in this review segregated into categories: Database Assisted, Feature Engineering, and Machine Learning}
\end{figure}

\clearpage
\bibliographystyle{IEEEtran}
\bibliography{MasterReference}

\begin{thebibliography}{10}
\providecommand{\url}[1]{#1}
\csname url@samestyle\endcsname
\providecommand{\newblock}{\relax}
\providecommand{\bibinfo}[2]{#2}
\providecommand{\BIBentrySTDinterwordspacing}{\spaceskip=0pt\relax}
\providecommand{\BIBentryALTinterwordstretchfactor}{4}
\providecommand{\BIBentryALTinterwordspacing}{\spaceskip=\fontdimen2\font plus
\BIBentryALTinterwordstretchfactor\fontdimen3\font minus
  \fontdimen4\font\relax}
\providecommand{\BIBforeignlanguage}[2]{{%
\expandafter\ifx\csname l@#1\endcsname\relax
\typeout{** WARNING: IEEEtran.bst: No hyphenation pattern has been}%
\typeout{** loaded for the language `#1'. Using the pattern for}%
\typeout{** the default language instead.}%
\else
\language=\csname l@#1\endcsname
\fi
#2}}
\providecommand{\BIBdecl}{\relax}
\BIBdecl

\bibitem{krebs1989ecological}
C.~J. Krebs \emph{et~al.}, ``Ecological methodology,'' Harper \& Row New York,
  Tech. Rep., 1989.

\bibitem{scheel2017second}
D.~Scheel, S.~Chancellor, M.~Hing, M.~Lawrence, S.~Linquist, and
  P.~Godfrey-Smith, ``A second site occupied by octopus tetricus at high
  densities, with notes on their ecology and behavior,'' \emph{Marine and
  Freshwater Behaviour and Physiology}, vol.~50, no.~4, pp. 285--291, 2017.

\bibitem{meek2013reliability}
P.~D. Meek, K.~Vernes, and G.~Falzon, ``On the reliability of expert
  identification of small-medium sized mammals from camera trap photos,''
  \emph{Wildlife Biology in Practice}, vol.~9, no.~2, pp. 1--19, 2013.

\bibitem{rowcliffe2008estimating}
J.~M. Rowcliffe, J.~Field, S.~T. Turvey, and C.~Carbone, ``Estimating animal
  density using camera traps without the need for individual recognition,''
  \emph{Journal of Applied Ecology}, vol.~45, no.~4, pp. 1228--1236, 2008.

\bibitem{burton2015wildlife}
A.~C. Burton, E.~Neilson, D.~Moreira, A.~Ladle, R.~Steenweg, J.~T. Fisher,
  E.~Bayne, and S.~Boutin, ``Wildlife camera trapping: a review and
  recommendations for linking surveys to ecological processes,'' \emph{Journal
  of Applied Ecology}, vol.~52, no.~3, pp. 675--685, 2015.

\bibitem{foster2012critique}
R.~J. Foster and B.~J. Harmsen, ``A critique of density estimation from
  camera-trap data,'' \emph{The Journal of Wildlife Management}, vol.~76,
  no.~2, pp. 224--236, 2012.

\bibitem{hiby2009tiger}
L.~Hiby, P.~Lovell, N.~Patil, N.~S. Kumar, A.~M. Gopalaswamy, and K.~U.
  Karanth, ``A tiger cannot change its stripes: using a three-dimensional model
  to match images of living tigers and tiger skins,'' \emph{Biology letters},
  pp. rsbl--2009, 2009.

\bibitem{zheng2015scalable}
L.~Zheng, L.~Shen, L.~Tian, S.~Wang, J.~Wang, and Q.~Tian, ``Scalable person
  re-identification: A benchmark,'' in \emph{Proceedings of the IEEE
  International Conference on Computer Vision}, 2015, pp. 1116--1124.

\bibitem{lecun2015deep}
Y.~LeCun, Y.~Bengio, and G.~Hinton, ``Deep learning,'' \emph{Nature}, vol. 521,
  no. 7553, pp. 436--444, 2015.

\bibitem{krizhevsky2012imagenet}
A.~Krizhevsky, I.~Sutskever, and G.~E. Hinton, ``Imagenet classification with
  deep convolutional neural networks,'' in \emph{Advances in neural information
  processing systems}, 2012, pp. 1097--1105.

\bibitem{lisanti2015person}
G.~Lisanti, I.~Masi, A.~D. Bagdanov, and A.~Del~Bimbo, ``Person
  re-identification by iterative re-weighted sparse ranking,'' \emph{IEEE
  transactions on pattern analysis and machine intelligence}, vol.~37, no.~8,
  pp. 1629--1642, 2015.

\bibitem{martinel2015re}
N.~Martinel, A.~Das, C.~Micheloni, and A.~K. Roy-Chowdhury, ``Re-identification
  in the function space of feature warps,'' \emph{IEEE transactions on pattern
  analysis and machine intelligence}, vol.~37, no.~8, pp. 1656--1669, 2015.

\bibitem{norouzzadehautomatically}
M.~S. Norouzzadeh, A.~Nguyen, M.~Kosmala, A.~Swanson, M.~Palmer, C.~Packer, and
  J.~Clune, ``Automatically identifying, counting, and describing wild animals
  in camera-trap images with deep learning,'' \emph{ArXiv:1703.05830v5}, 2017.

\bibitem{schneider2018deep}
S.~Schneider, G.~Taylor, and S.~Kremer, ``Deep learning object detection
  methods for ecological camera trap data,'' \emph{Conference on Computer and
  Robot Vision}, to appear.

\bibitem{carter2014automated}
S.~J. Carter, I.~P. Bell, J.~J. Miller, and P.~P. Gash, ``Automated marine
  turtle photograph identification using artificial neural networks, with
  application to green turtles,'' \emph{Journal of experimental marine biology
  and ecology}, vol. 452, pp. 105--110, 2014.

\bibitem{freytag2016chimpanzee}
A.~Freytag, E.~Rodner, M.~Simon, A.~Loos, H.~S. K{\"u}hl, and J.~Denzler,
  ``Chimpanzee faces in the wild: Log-euclidean cnns for predicting identities
  and attributes of primates,'' in \emph{German Conference on Pattern
  Recognition}.\hskip 1em plus 0.5em minus 0.4em\relax Springer, 2016, pp.
  51--63.

\bibitem{brust2017towards}
C.-A. Brust, T.~Burghardt, M.~Groenenberg, C.~K{\"a}ding, H.~S. K{\"u}hl, M.~L.
  Manguette, and J.~Denzler, ``Towards automated visual monitoring of
  individual gorillas in the wild,'' in \emph{Proceedings of the IEEE
  Conference on Computer Vision and Pattern Recognition}, 2017, pp. 2820--2830.

\bibitem{loos2013automated}
A.~Loos and A.~Ernst, ``An automated chimpanzee identification system using
  face detection and recognition,'' \emph{EURASIP Journal on Image and Video
  Processing}, vol. 2013, no.~1, p.~49, 2013.

\bibitem{whitehead1990computer}
H.~Whitehead, ``Computer assisted individual identification of sperm whale
  flukes,'' \emph{Reports of the International Whaling Commission}, vol.~12,
  pp. 71--77, 1990.

\bibitem{mizroch1990computer}
S.~A. Mizroch, J.~A. Beard, and M.~Lynde, ``Computer assisted
  photo-identification of humpback whales,'' \emph{Report of the International
  Whaling Commission}, vol.~12, pp. 63--70, 1990.

\bibitem{hiby1990computer}
L.~Hiby and P.~Lovell, ``Computer aided matching of natural markings: a
  prototype system for grey seals,'' \emph{Report of the International Whaling
  Commission}, vol.~12, pp. 57--61, 1990.

\bibitem{kagglehumpbackreid}
``Humpback whale identification challenge,''
  \url{https://www.kaggle.com/c/whale-categorization-playground}, accessed:
  2018-05-15.

\bibitem{o1998identification}
G.~O’Corry-Crowe, ``Identification of individual sperm whales by wavelet
  transform of the trailing edge of the flukes,'' \emph{Marine Mammal Science},
  vol.~14, no.~1, pp. 143--145, 1998.

\bibitem{kelly2001computer}
M.~J. Kelly, ``Computer-aided photograph matching in studies using individual
  identification: an example from {S}erengeti cheetahs,'' \emph{Journal of
  Mammalogy}, vol.~82, no.~2, pp. 440--449, 2001.

\bibitem{hillman2003computer}
G.~Hillman, B.~Wursig, G.~Gailey, N.~Kehtarnavaz, A.~Drobyshevsky, B.~Araabi,
  H.~Tagare, and D.~Weller, ``Computer-assisted photo-identification of
  individual marine vertebrates: a multi-species system,'' \emph{Aquatic
  Mammals}, vol.~29, no.~1, pp. 117--123, 2003.

\bibitem{ravela2004recognizing}
S.~Ravela and L.~Gamble, ``On recognizing individual salamanders,'' in
  \emph{Proceedings of Asian Conference on Computer Vision, Ki-Sang Hong and
  Zhengyou Zhang, Ed. Jeju, Korea}, 2004, pp. 742--747.

\bibitem{groth1986pattern}
E.~J. Groth, ``A pattern-matching algorithm for two-dimensional coordinate
  lists,'' \emph{The astronomical journal}, vol.~91, pp. 1244--1248, 1986.

\bibitem{arzoumanian2005astronomical}
Z.~Arzoumanian, J.~Holmberg, and B.~Norman, ``An astronomical pattern-matching
  algorithm for computer-aided identification of whale sharks \emph{{R}hincodon
  typus},'' \emph{Journal of Applied Ecology}, vol.~42, no.~6, pp. 999--1011,
  2005.

\bibitem{ardovini2008identifying}
A.~Ardovini, L.~Cinque, and E.~Sangineto, ``Identifying elephant photos by
  multi-curve matching,'' \emph{Pattern Recognition}, vol.~41, no.~6, pp.
  1867--1877, 2008.

\bibitem{burghardt2007fully}
T.~Burghardt, P.~Barham, N.~Campbell, I.~Cuthill, R.~Sherley, and T.~Leshoro,
  ``A fully automated computer vision system for the biometric identification
  of {A}frican penguins \emph{{S}pheniscus demersus} on {R}obben {I}sland,'' in
  \emph{Abstracts of Oral and Poster Presentations, 6th International Penguin
  Conference. Birds Tasmania, Hobart}, 2007.

\bibitem{sherley2010spotting}
R.~B. Sherley, T.~Burghardt, P.~J. Barham, N.~Campbell, and I.~C. Cuthill,
  ``Spotting the difference: towards fully-automated population monitoring of
  african penguins spheniscus demersus,'' \emph{Endangered Species Research},
  vol.~11, no.~2, pp. 101--111, 2010.

\bibitem{town2013manta}
C.~Town, A.~Marshall, and N.~Sethasathien, ``Manta {M}atcher: automated
  photographic identification of manta rays using keypoint features,''
  \emph{Ecology and evolution}, vol.~3, no.~7, pp. 1902--1914, 2013.

\bibitem{hearst1998support}
M.~A. Hearst, S.~T. Dumais, E.~Osuna, J.~Platt, and B.~Scholkopf, ``Support
  vector machines,'' \emph{IEEE Intelligent Systems and their applications},
  vol.~13, no.~4, pp. 18--28, 1998.

\bibitem{hughes2017automated}
B.~Hughes and T.~Burghardt, ``Automated visual fin identification of individual
  great white sharks,'' \emph{International Journal of Computer Vision}, vol.
  122, no.~3, pp. 542--557, 2017.

\bibitem{mcculloch1943logical}
W.~S. McCulloch and W.~Pitts, ``A logical calculus of the ideas immanent in
  nervous activity,'' \emph{The bulletin of mathematical biophysics}, vol.~5,
  no.~4, pp. 115--133, 1943.

\bibitem{rumelhart1986learning}
D.~E. Rumelhart, G.~E. Hinton, and R.~J. Williams, ``Learning representations
  by back-propagating errors,'' \emph{nature}, vol. 323, no. 6088, p. 533,
  1986.

\bibitem{hinton2006fast}
G.~E. Hinton, S.~Osindero, and Y.-W. Teh, ``A fast learning algorithm for deep
  belief nets,'' \emph{Neural computation}, vol.~18, no.~7, pp. 1527--1554,
  2006.

\bibitem{goodfellow2016deep}
I.~Goodfellow, Y.~Bengio, and A.~Courville, \emph{Deep learning}.\hskip 1em
  plus 0.5em minus 0.4em\relax MIT press, 2016.

\bibitem{simonyan2014very}
K.~Simonyan and A.~Zisserman, ``Very deep convolutional networks for
  large-scale image recognition,'' \emph{arXiv preprint arXiv:1409.1556}, 2014.

\bibitem{szegedy2015going}
C.~Szegedy, W.~Liu, Y.~Jia, P.~Sermanet, S.~Reed, D.~Anguelov, D.~Erhan,
  V.~Vanhoucke, and A.~Rabinovich, ``Going deeper with convolutions,'' in
  \emph{Proceedings of the IEEE conference on computer vision and pattern
  recognition}, 2015, pp. 1--9.

\bibitem{he2016deep}
K.~He, X.~Zhang, S.~Ren, and J.~Sun, ``Deep residual learning for image
  recognition,'' in \emph{Proceedings of the IEEE conference on computer vision
  and pattern recognition}, 2016, pp. 770--778.

\bibitem{hornik1991approximation}
K.~Hornik, ``Approximation capabilities of multilayer feedforward networks,''
  \emph{Neural networks}, vol.~4, no.~2, pp. 251--257, 1991.

\bibitem{swanson2015snapshot}
A.~Swanson, M.~Kosmala, C.~Lintott, R.~Simpson, A.~Smith, and C.~Packer,
  ``Snapshot serengeti, high-frequency annotated camera trap images of 40
  mammalian species in an {A}frican savanna,'' \emph{Scientific data}, vol.~2,
  2015.

\bibitem{fukushima1979neural}
K.~Fukushima, ``Neural network model for a mechanism of pattern recognition
  unaffected by shift in position- neocognitron,'' \emph{Electron. \& Commun.
  Japan}, vol.~62, no.~10, pp. 11--18, 1979.

\bibitem{iandola2014densenet}
F.~Iandola, M.~Moskewicz, S.~Karayev, R.~Girshick, T.~Darrell, and K.~Keutzer,
  ``Densenet: Implementing efficient convnet descriptor pyramids,'' \emph{arXiv
  preprint arXiv:1404.1869}, 2014.

\bibitem{sabour2017dynamic}
S.~Sabour, N.~Frosst, and G.~E. Hinton, ``Dynamic routing between capsules,''
  in \emph{Advances in Neural Information Processing Systems}, 2017, pp.
  3856--3866.

\bibitem{pan2010survey}
S.~J. Pan and Q.~Yang, ``A survey on transfer learning,'' \emph{IEEE
  Transactions on knowledge and data engineering}, vol.~22, no.~10, pp.
  1345--1359, 2010.

\bibitem{rodner2015fine}
E.~Rodner, M.~Simon, G.~Brehm, S.~Pietsch, J.~W. W{\"a}gele, and J.~Denzler,
  ``Fine-grained recognition datasets for biodiversity analysis,'' \emph{arXiv
  preprint arXiv:1507.00913}, 2015.

\bibitem{rodner2016fine}
E.~Rodner, M.~Simon, R.~B. Fisher, and J.~Denzler, ``Fine-grained recognition
  in the noisy wild: Sensitivity analysis of convolutional neural networks
  approaches,'' \emph{arXiv preprint arXiv:1610.06756}, 2016.

\bibitem{souri2015fast}
Y.~Souri and S.~Kasaei, ``Fast bird part localization for fine-grained
  categorization,'' in \emph{Proc. 3rd Workshop Fine-Grained Vis.
  Categorization (FGVC3) CVPR}, 2015.

\bibitem{ren2017faster}
S.~Ren, K.~He, R.~Girshick, and J.~Sun, ``Faster {R-CNN}: Towards real-time
  object detection with region proposal networks,'' \emph{IEEE transactions on
  pattern analysis and machine intelligence}, vol.~39, no.~6, pp. 1137--1149,
  2017.

\bibitem{redmon2016you}
J.~Redmon, S.~Divvala, R.~Girshick, and A.~Farhadi, ``You only look once:
  Unified, real-time object detection,'' in \emph{Proceedings of the IEEE
  Conference on Computer Vision and Pattern Recognition}, 2016, pp. 779--788.

\bibitem{liu2016ssd}
W.~Liu, D.~Anguelov, D.~Erhan, C.~Szegedy, S.~Reed, C.-Y. Fu, and A.~C. Berg,
  ``Ssd: Single shot multibox detector,'' in \emph{European conference on
  computer vision}.\hskip 1em plus 0.5em minus 0.4em\relax Springer, 2016, pp.
  21--37.

\bibitem{nowozin2014optimal}
S.~Nowozin, ``Optimal decisions from probabilistic models: the
  intersection-over-union case,'' in \emph{Proceedings of the IEEE Conference
  on Computer Vision and Pattern Recognition}, 2014, pp. 548--555.

\bibitem{taigman2014deepface}
Y.~Taigman, M.~Yang, M.~Ranzato, and L.~Wolf, ``Deepface: Closing the gap to
  human-level performance in face verification,'' in \emph{Proceedings of the
  IEEE conference on computer vision and pattern recognition}, 2014, pp.
  1701--1708.

\bibitem{bromley1994signature}
J.~Bromley, I.~Guyon, Y.~LeCun, E.~S{\"a}ckinger, and R.~Shah, ``Signature
  verification using a "siamese" time delay neural network,'' in \emph{Advances
  in Neural Information Processing Systems}, 1994, pp. 737--744.

\bibitem{schroff2015facenet}
F.~Schroff, D.~Kalenichenko, and J.~Philbin, ``Facenet: A unified embedding for
  face recognition and clustering,'' in \emph{Proceedings of the IEEE
  conference on computer vision and pattern recognition}, 2015, pp. 815--823.

\bibitem{Koerschens18:Elephants}
M.~Körschens, B.~Barz, and J.~Denzler, ``Towards automatic identification of
  elephants in the wild,'' in \emph{AI for Wildlife Conservation Workshop
  (AIWC)}, 2018.

\bibitem{deb2018face}
D.~Deb, S.~Wiper, A.~Russo, S.~Gong, Y.~Shi, C.~Tymoszek, and A.~Jain, ``Face
  recognition: Primates in the wild,'' \emph{arXiv preprint arXiv:1804.08790},
  2018.

\bibitem{zhou2017brief}
Z.-H. Zhou, ``A brief introduction to weakly supervised learning,''
  \emph{National Science Review}, vol.~5, no.~1, pp. 44--53, 2017.

\bibitem{simpson2014zooniverse}
R.~Simpson, K.~R. Page, and D.~De~Roure, ``Zooniverse: observing the world's
  largest citizen science platform,'' in \emph{Proceedings of the 23rd
  international conference on world wide web}.\hskip 1em plus 0.5em minus
  0.4em\relax ACM, 2014, pp. 1049--1054.

\bibitem{holzinger2016interactive}
A.~Holzinger, ``Interactive machine learning for health informatics: when do we
  need the human-in-the-loop?'' \emph{Brain Informatics}, vol.~3, no.~2, pp.
  119--131, 2016.

\bibitem{kaggleiWildCam2018}
``i{W}ildcam 2018 camera trap challenge,''
  \url{https://www.kaggle.com/c/iwildcam2018}, accessed: 2018-07-11.

\end{thebibliography}

\end{document}